\documentclass[runningheads]{llncs}
\usepackage{graphicx}
\usepackage{booktabs}
\usepackage{multirow}
\usepackage{makecell}
\usepackage{marvosym}
\usepackage{amsmath,amssymb}
\usepackage{threeparttable}
\begin{document}
\title{CatPortrait: Taming Video Generative Models for Portrait Animation via In-context Conditioning}
%
%
\newcommand{\corrauthor}{%
  \textsuperscript{(\raisebox{-0.1ex}{\scalebox{1.3}{\Letter}})}}
\author{Ziqi Ni\inst{1,2}\and
Zhou Fang\inst{1,2} \and
Shijian Wang\inst{1} \and
Rui Li\inst{2} \and
Yi Zhou\inst{1,2}\corrauthor}
%
\authorrunning{F. Author et al.}
%
\institute{
School of Computer Science and Engineering, Southeast University, Nanjing, China
\and
Key Laboratory of New Generation Artificial Intelligence Technology and Its Interdisciplinary Applications, Ministry of Education, China\\
\email{yizhou.szcn@gmail.com}
\and
University of Science and Technology of China, Hefei, China
}
\maketitle              
\begin{abstract}
Recent video generative models have shown strong capability in synthesizing high-quality and temporally coherent videos. Yet, adapting them for controllable portrait animation remains nontrivial, as existing methods typically rely on architectural changes or extra conditioning modules, increasing complexity and reducing compatibility with pretrained models.
We present \textbf{CatPortrait}, a simple and effective framework that enables controllable portrait animation via 3D face guidance as in-context conditioning. CatPortrait preserves the original video generation backbone and injects structured 3D facial guidance into the input space through spatial concatenation, allowing pretrained image-to-video models to animate portraits without sacrificing their pretrained priors. CatPortrait supports both video-driven and audio-driven portrait animation within a unified framework. Extensive experiments demonstrate that CatPortrait achieves competitive or superior results across both settings.

\end{abstract}

\section{Introduction}
Portrait animation, also known as talking head generation, aims to synthesize realistic facial videos driven by audio or reference motion while preserving identity and visual fidelity. It enables applications such as digital avatars, virtual meetings, and content creation, yet remains challenging to achieve both high quality and controllability.

Recent video foundation models~\cite{wan2025wan,kong2024hunyuanvideo,ni2026seeing,li2026learning,li2026reward,liang2026teleboost,li2025growing,liang2026integrating} can generate high-resolution and temporally coherent videos thanks to large-scale pretraining. However, directly adapting them to portrait animation is non-trivial: unconditional generation lacks precise motion control, while existing controllable methods~\cite{hallo,xportrait,aniportrait,hunyuanportrait} often introduce adapters, cross-attention branches, or modality-specific networks, increasing complexity and partially reducing the generality of pretrained models.

We revisit portrait animation from the perspective of in-context conditioning (ICC) and propose \textbf{CatPortrait}, which controls pretrained image-to-video models by spatially concatenating texture-free 3D facial guidance in the input space. Leveraging the strong reconstruction capability of modern VAEs~\cite{wan2025wan} and the disentangled representation of 3DMMs~\cite{3dmm}, CatPortrait enables controllable portrait animation without modifying the video backbone.

CatPortrait unifies video-driven and audio-driven animation in a single framework. For video-driven animation, identity coefficients are extracted from the reference portrait and motion coefficients from a driving video; for audio-driven animation, a lightweight audio-to-FLAME model predicts motion from speech. The coefficients are rendered into a guidance video and concatenated with the visual input to synthesize the final talking-head video, introducing no additional parameters to the video generator.
Our contributions are summarized as follows:
\begin{itemize}
\item We propose CatPortrait, an ICC-based framework that adapts pretrained video generative models for controllable portrait animation via spatial concatenation without task-specific control modules.
\item We instantiate ICC with a disentangled 3D face representation, yielding a unified framework for both audio-driven and video-driven portrait animation with compositional control.
\item Experiments demonstrate competitive performance on standard benchmarks with strong robustness and generalization.
\end{itemize}

\section{Related Work}
\label{sec:related}
\noindent\textbf{Talking Head Generation.}
Talking head generation aims to synthesize realistic facial videos driven by speech or reference motion. Existing methods~\cite{wav2lip,aniportrait,xportrait,hunyuanportrait,freak,dmac} can be broadly categorized into audio-driven and video-driven approaches.

Audio-driven methods generate facial motion, particularly lip movements, from speech. Early works predict explicit facial representations for rendering~\cite{makeittalk,iplap} or formulate the task as audio-guided face editing~\cite{wav2lip,difftalk}. More recent diffusion-based methods~\cite{hallo,cui2024hallo2} improve visual quality and temporal coherence, but often rely on specialized audio-conditioning modules, identity-reference designs, or multi-stage training.

Video-driven methods, also known as facial reenactment, transfer pose and expressions from a driving video to a target portrait. Earlier approaches~\cite{dagan,facavid2vid} mainly use landmarks, learned keypoints, or dense motion fields, while recent diffusion-based methods~\cite{xportrait,hunyuanportrait,wei2025magicportrait} introduce richer structural or visual controls. However, they still commonly depend on task-specific warping schemes, control branches, or reference modules.

A related line of work seeks to unify audio-driven and video-driven portrait animation with shared intermediate representations~\cite{aniportrait,difftalker}. In particular, 3DMMs provide disentangled representations of identity, expression, and pose~\cite{3dsurvey}. Nevertheless, existing methods usually inject them through dedicated adapters or task-specific control architectures, requiring non-trivial modifications to the generative backbone.

\noindent\textbf{In-Context Conditioning.}
Recently, in-context conditioning (ICC) has emerged as a promising paradigm for controllable video generation and editing~\cite{ju2025fulldit,video_icl,ju2025editverse,cai2025videocanvas,catvton}. Instead of introducing task-specific adapters or auxiliary control branches, ICC-based methods place reference content and control signals directly in context, allowing pretrained video diffusion transformers to model the target output jointly with the provided conditions. Early evidence shows that video diffusion transformers can exhibit in-context generation capability under simple spatial or temporal concatenation with minimal tuning. More recent works further unify diverse control signals as a single token sequence processed by full self-attention, enabling multi-task video generation and editing without specialized architectural modifications. This line has also been extended to broader multimodal settings and finer spatio-temporal control, such as arbitrary patch-based video completion~\cite{spatialtemp-freq,neuroevolution,vlm}.

Our work is inspired by this emerging ICC perspective, but differs from prior token-level or patch-level formulations in a fundamental way. Rather than targeting generic video editing or arbitrary spatio-temporal completion, CatPortrait instantiates ICC for portrait animation using a structured 3D facial prior. Specifically, we render disentangled 3DMM-based facial motion into a guidance video and inject it through spatial concatenation in the input space. This design preserves the pretrained video backbone while leveraging explicit 3D identity-motion factorization, making the framework naturally suitable for both audio-driven and video-driven portrait animation.

\section{Method}
\label{sec:method}
\subsection{Preliminaries}
\noindent\textbf{Video Generation Model.}
Our study is based on Wan~\cite{wan2025wan}, a latent video diffusion framework composed of (i) a causal 3D VAE that compresses videos along spatial and temporal dimensions, and (ii) a Transformer-based diffusion model (DiT)~\cite{dit}. The generative process follows Rectified Flow~\cite{rectifiedflow}, where the forward trajectory is defined as a linear interpolation between clean data and Gaussian noise:
\begin{equation}
\label{flowmatching_eq}
    z_t = (1-t)z_0 + t\epsilon, \quad \epsilon \sim \mathcal{N}(0, I),
\end{equation}
where $t$ denotes the continuous timestep. The model is trained with Conditional Flow Matching~\cite{flowmatching}, with objective
\begin{equation}
    \mathcal{L}_{\mathrm{CFM}} = \mathbb{E}_{t,\,p_t(z, \epsilon),\,p(\epsilon)} 
    \left\| v_{\Theta}(z_t, t) - u_t(z_0 \mid \epsilon) \right\|_2^2 .
\end{equation}

\noindent\textbf{3D Morphable Face Model.}
We use FLAME~\cite{flame} as our 3D Morphable Face Model for its disentangled parameterization of facial identity and motion. Given identity coefficients $\beta$, expression coefficients $\psi$, and pose parameters $\theta$, FLAME generates a posed mesh through blendshapes and linear blend skinning:
\begin{equation}
    \mathbf{V}(\beta,\psi,\theta)
    = \mathrm{LBS}\Big(\bar{\mathbf{V}} + \mathbf{B}_{\mathrm{sh}}\beta + \mathbf{B}_{\mathrm{exp}}\psi,\,
    \mathbf{J}(\beta),\,\theta,\,\mathcal{W}\Big),
\end{equation}
where $\bar{\mathbf{V}}$ is the template mesh, $\mathbf{B}_{\mathrm{sh}}$ and $\mathbf{B}_{\mathrm{exp}}$ are the shape and expression bases, $\mathbf{J}(\cdot)$ denotes joints, and $\mathcal{W}$ are skinning weights. For convenience, we write the resulting facial shape as
\begin{equation}
\mathbf{S} = \mathbf{S}(\beta, \psi, \theta).
\end{equation}
This factorization explicitly separates identity from motion, which is crucial for controllable portrait animation.

\begin{figure*}
    \centering
    \includegraphics[width=1\linewidth]{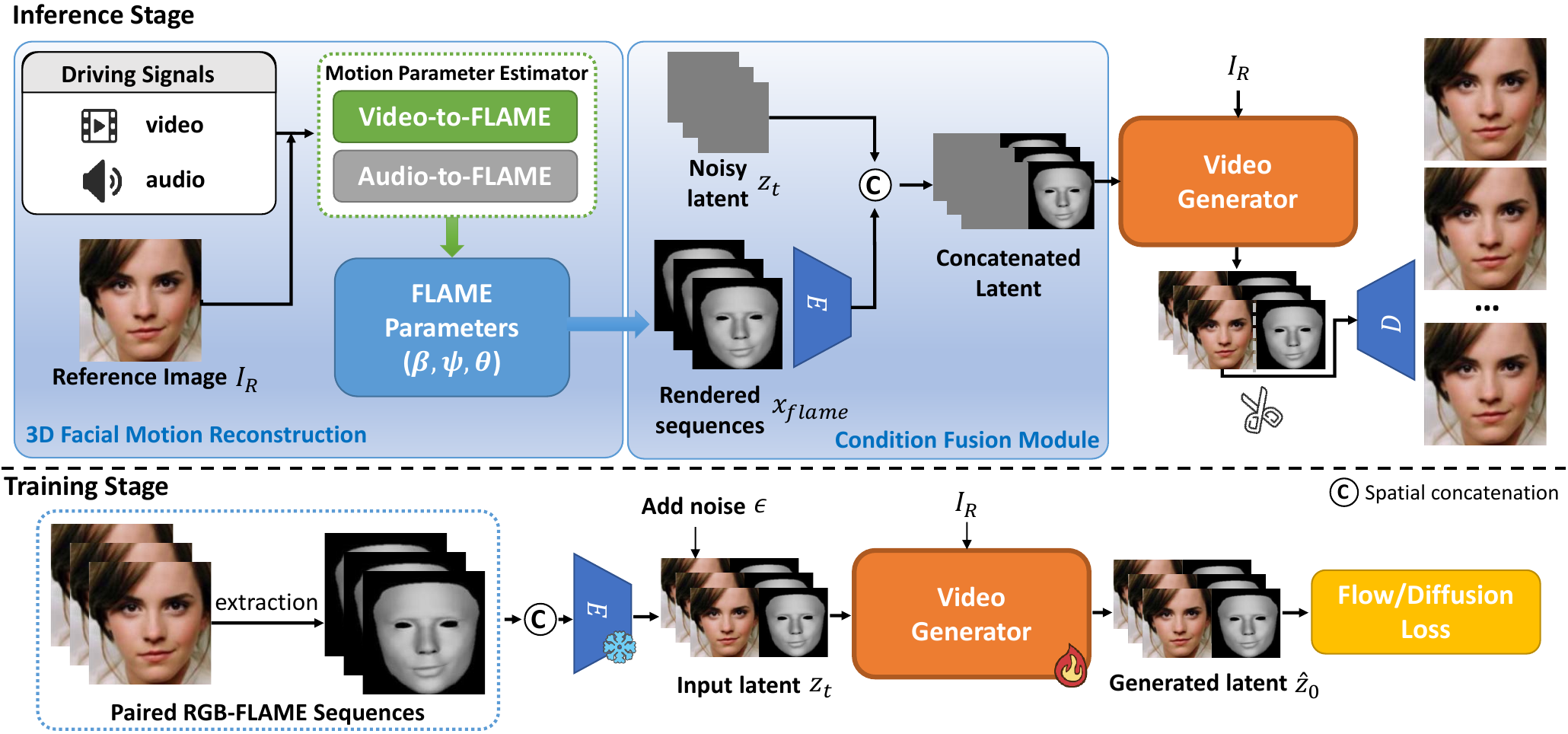}
    \caption{\textbf{Overview of the proposed framework.} During inference, the model takes a reference image $I_R$ and driving signals as inputs. A motion parameter estimator converts the driving signals into FLAME parameters $(\beta,\psi,\theta)$. The rendered FLAME sequence is then encoded by $E$ and spatially concatenated with the noisy latent $z_t$ to form a conditioned latent, which, together with the identity information from $I_R$, is fed into the video generator to synthesize the output video. During training, paired sequences are used to construct the input, and the generator is finetuned.}
    \label{fig:overview}
\end{figure*}

\subsection{Overview}
We build CatPortrait on top of a pretrained image-to-video model and adapt it for portrait animation without modifying its architecture. As illustrated in Fig.~\ref{fig:overview}, given a reference portrait and a driving signal, we first estimate FLAME coefficients, where the identity is determined by the reference portrait and the facial motion is driven by either a video or speech input. Specifically, these coefficients are parameterized as $(\beta,\psi,\theta)$, where $\beta$ controls identity, while $\psi$ and $\theta$ encode expression and pose, respectively. We then render the estimated coefficients into a texture-free 3D face guidance clip that explicitly captures the target facial dynamics. The rendered guidance is spatially combined with the appearance context to form a composite clip, which is encoded by the pretrained video VAE into the latent space. During denoising, the guidance region is clamped as known context, and the model synthesizes the portrait region conditioned on the rendered 3D motion. Finally, the animated result is obtained by cropping the generated portrait half from the synthesized video.

\subsection{3D Face Guidance as In-Context Conditioning}
Specifically, for each sample, we represent facial identity with the FLAME identity coefficient vector $\beta$ estimated from the reference portrait $I_R$, and represent facial motion with frame-wise expression and pose parameters $\{(\psi^t,\theta^t)\}_{t=1}^{T}$. In the video-driven setting, $(\psi^t,\theta^t)$ are extracted from the driving video using a video-to-FLAME estimator~\cite{smirk}, while in the audio-driven setting they are predicted from speech by an audio-to-FLAME model. We combine them into a complete FLAME trajectory $\{(\beta,\psi^t,\theta^t)\}_{t=1}^{T}$.

Given the FLAME trajectory, we render each frame into a texture-free guidance image $G_t$ that preserves facial geometry and dynamics, including pose, expression, and identity-related shape cues, while discarding photorealistic texture. We use a fixed camera and rendering setup, and represent each rendered face as a geometry-centric RGB image, so that the resulting guidance clip $G_{1:T}$ can be directly encoded by the pretrained video VAE.

Let $A_{1:T}$ denote the portrait appearance clip used during training. We form a composite clip by concatenating the portrait frame and the rendered guidance frame along the spatial width:
\begin{equation}
    C_t = \mathrm{Concat}_{w}\big(A_t, G_t\big), \quad t=1,\ldots,T,
\end{equation}
so that each frame $C_t \in \mathbb{R}^{H \times 2W \times 3}$ contains both appearance and motion context. The full composite video is denoted by $C_{1:T}$.

During training, $A_{1:T}$ is the ground-truth portrait video clip, and the composite clip $C_{1:T}$ is encoded by the pretrained video VAE into the clean latent $z_0=\mathcal{E}(C_{1:T})$. During inference, generation starts from a noise latent in the portrait region, while the rendered guidance region is injected as observed context throughout denoising. The reference portrait $I_R$ is provided separately as the appearance condition to preserve the target identity.

We treat the guidance region as observed context and synthesize the portrait region conditioned on it. Let $\mathcal{M}$ be a binary spatial mask in latent space indicating the observed region. In our setting, $\mathcal{M}$ covers the entire guidance half for all frames and can optionally include the first portrait frame to further anchor identity. We sample a timestep $t$ and noise $\epsilon$, and construct the noisy latent $z_t$ following Eq.~\ref{flowmatching_eq}. To impose in-context conditioning, we replace the observed region in the noisy latent with its clean counterpart:
\begin{equation}
    \tilde{z}_t = \mathcal{M}\odot z_0 + (1-\mathcal{M})\odot z_t.
\end{equation}
The original video backbone is then trained on $\tilde{z}_t$ with the standard denoising objective, without any architectural modification. At inference time, we apply the same operation at every denoising step so that the generated portrait remains consistent with the provided 3D guidance.

This in-context conditioning design is effective for two reasons. First, the rendered FLAME guidance provides an explicit and spatially aligned representation of facial motion, making the control signal easier to interpret than implicit latent conditions. Second, by keeping the guidance region observable throughout denoising, the pretrained video generator can reuse its native inpainting-style generation capability to synthesize the portrait region under strong structural constraints, without requiring architectural changes or additional motion-specific modules.

\subsection{Audio-to-FLAME Module}
\label{sec:audio2flame}
For audio-driven portrait animation, we require a mapping from speech to a time-varying facial motion trajectory. CatPortrait is agnostic to the specific speech-to-3D-face predictor; in principle, any existing model that predicts FLAME motion parameters or compatible intermediate representations can be used as the motion provider~\cite{fan2022faceformer}. In this work, we instantiate this module as a lightweight audio-to-FLAME predictor for temporally stable motion generation.

Given an audio signal $a$, we first extract frame-aligned audio features $\{f_t\}_{t=1}^{T}$ that match the target video FPS. The module then predicts frame-wise motion parameters
\begin{equation}
    \{(\hat{\psi}^{t},\hat{\theta}^{t})\}_{t=1}^{T}
    = \mathcal{A}\big(\{f_t\}_{t=1}^{T}\big),
\end{equation}
while the identity coefficient $\beta$ is estimated from the reference portrait and kept fixed over time. To improve temporal consistency, we implement $\mathcal{A}(\cdot)$ as a diffusion-style temporal generator in FLAME coefficient space~\cite{mdm}. Specifically, an audio encoder~\cite{hubert} first maps the speech signal into feature embeddings, and a lightweight Transformer-based denoiser is trained to recover clean motion trajectories from noisy ones conditioned on the audio embeddings and the diffusion timestep.

For supervision, we obtain pseudo ground-truth FLAME parameters from real portrait videos using an off-the-shelf FLAME regressor~\cite{smirk}, which enables training on large-scale in-the-wild talking videos without motion-capture annotations.

\section{Experiments}
\label{sec:experiment}
\subsection{Experimental setup}
\noindent\textbf{Implementation Details}
CatPortrait is built on the pretrained Wan2.1-1.3B-I2V with the backbone unchanged. We follow Wan's original training objective and default optimization settings, and only modify the input by spatially concatenating the rendered 3D face guidance with the portrait video. We fine-tune for 5 epochs on 2$\times$NVIDIA A6000 GPUs with batch size 1 and learning rate $1\times10^{-5}$, while keeping the pretrained video VAE frozen.

\noindent\textbf{Datasets}
Following prior works~\cite{dinet,difftalk}, we use HDTF~\cite{hdtf} and MEAD~\cite{wang2020mead} for training and evaluation with an 80/20 train/test split. All videos are processed to $256\times256$. For HDTF, we filter low-resolution clips and samples with unreliable face detection; we then detect and crop face regions frame-by-frame to obtain processed videos for 3DMM extraction and model training.

\noindent\textbf{Compared Baselines}
We compare with recent talking portrait methods. For audio-driven animation, we include Hallo~\cite{hallo}, Hallo2~\cite{cui2024hallo2}, SadTalker~\cite{sadtalker}, and AniPortrait~\cite{aniportrait}. For video-driven animation, we include AniPortrait~\cite{aniportrait}, X-Portrait~\cite{xportrait}, and HunyuanPortrait~\cite{hunyuanportrait}. We use official checkpoints or public implementations when available and evaluate all methods under the same protocol.

\noindent\textbf{Evaluation Metrics}
For audio-driven animation, we report FID and FVD~\cite{fvd} for visual/spatio-temporal quality, and SyncNet-based LSE-D/LSE-C~\cite{wav2lip} plus M-LMD~\cite{edtalk} for lip-sync. For video-driven animation, we report self-reenactment using FID/FVD and LMD~\cite{lmd}/LPIPS~\cite{lpips} and cross-reenactment additionally reports AED and AKD~\cite{akd} following~\cite{dagan,xportrait}, as well as identity similarity (ID Sim.)~\cite{deng2019arcface} and Image Quality~\cite{hyperIQA}.

\subsection{Quantitative Evaluation}
\begin{table*}[htbp]
\caption{Quantitative comparison on audio-driven talking head generation.
}
\centering
\begin{tabular}{|c|c|c|c|c|c|}
\hline
Method &\textbf{FID} &\textbf{FVD}& \textbf{LSE-D}& \textbf{LSE-C} & \textbf{M-LMD}\\
\hline
Hallo\cite{hallo}& 82.7286  & 132.09  &  8.0047  &  7.0980  &   8.0245 \\
Hallo2\cite{cui2024hallo2}&  147.1788 & 130.65  &  8.2012  &  7.067  &  6.5618  \\
SadTalker\cite{sadtalker}& 685.7087  & 305.59  & 7.9159   & 6.8427  & 3.6368   \\
AniPortrait\cite{aniportrait}& 138.636 & 414.77  & 12.6616 & 4.1945   &  2.3688  \\
Ours & \textbf{81.1032} & \textbf{113.53}&  8.2837& 6.8823 & \textbf{1.632}\\
\hline
\end{tabular}
\label{quan_aud}
\end{table*}
\noindent\textbf{Audio-driven talking head generation.}
Table~\ref{quan_aud} presents the quantitative results on audio-driven talking head generation. CatPortrait achieves the best performance on FID, FVD, and M-LMD among all compared methods, indicating strong visual quality, temporal coherence, and mouth-motion accuracy. In particular, the gains on FID and FVD suggest that the proposed in-context 3D guidance effectively leverages the pretrained video prior for high-quality and coherent video synthesis, while the improvement on M-LMD shows more accurate facial motion under speech control.

CatPortrait is not the best on the SyncNet-based metrics LSE-D and LSE-C. This is expected, since our method does not employ specialized lip-sync modules or explicit synchronization losses commonly used in methods optimized primarily for speech-lip alignment. Nevertheless, the overall results show that CatPortrait achieves a favorable balance between perceptual quality, temporal coherence, and motion accuracy, highlighting the effectiveness of structured 3D facial motion as an intermediate control signal for audio-driven portrait animation.
\begin{table*}[htbp]
\caption{Quantitative comparison on video-driven portrait reenactment under self-reenactment and cross-reenactment settings.}
\centering
\resizebox{0.98\textwidth}{!}{
\begin{tabular}{|c|c|c|c|c|c|c|c|c|}
\hline
\textbf{Method}&\multicolumn{4}{|c|}{\textbf{Self-Reenactment}}&\multicolumn{4}{|c|}{\textbf{Cross-Reenactment}}\\
\cline{2-9} 
 & FID & FVD & LMD & LPIPS & AED & AKD &ID Sim.& Image Qual. \\
\hline
AniPortrait\cite{aniportrait}& 155.190 & 413.48 &2.8264& 0.2105 &1.1301 &13.5388 & 0.5753 & 30.5514 \\
X-Portrait\cite{xportrait}&89.47 & 149.61 &11.599 &0.0985 & 0.9661 &7.087 & 0.8445 & 35.4013\\
HunyuanPortrait\cite{hunyuanportrait}& 82.609& 108.02 & 1.003& 0.0666& 0.9585& 6.9574 &0.8290 &  36.4791 \\
Ours & \underline{88.79} & \textbf{104.99} & \underline{1.503} & \underline{0.0757} & \textbf{0.9563} & \underline{6.9923} & 0.8025 & \textbf{37.5668} \\
\hline
\end{tabular}
}
\label{quan_vid}
\end{table*}

\noindent\textbf{Video-driven reenactment.}
Table~\ref{quan_vid} reports the quantitative results on video-driven portrait reenactment under both self-reenactment and cross-reenactment settings. In the self-reenactment setting, CatPortrait achieves the best FVD and the second-best results on FID, LMD, and LPIPS. Although HunyuanPortrait remains the strongest baseline overall in this setting, CatPortrait consistently performs close to it while outperforming the other compared methods, demonstrating strong visual quality, temporal coherence, and motion fidelity without task-specific reenactment modules.

In the more challenging cross-reenactment setting, CatPortrait achieves the best AED and Image Quality, and the second-best AKD. These results indicate that our method transfers facial motion faithfully across identities while maintaining strong visual quality. The gains on AED and AKD support the effectiveness of the proposed 3D guidance in decoupling motion from source appearance and recomposing it onto a target identity. Meanwhile, the best Image Quality score suggests that CatPortrait better preserves the generative fidelity of the pretrained video model under cross-identity control. CatPortrait does not achieve the best identity similarity, indicating that preserving fine-grained identity cues under large cross-identity motion transfer remains an important direction for improvement.

\subsection{Qualitative Evaluation}
\begin{figure}
    \centering
    \includegraphics[width=1\linewidth]{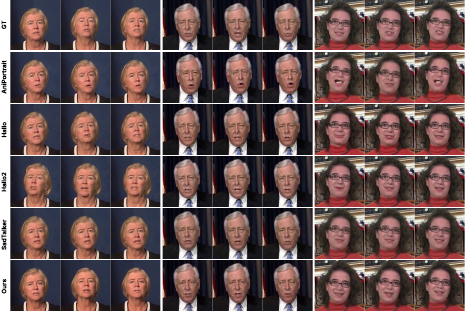}
    \caption{Qualitative comparison on audio-driven talking head generation.}
    \label{fig:aud_res}
\end{figure}
\noindent\textbf{Audio-driven talking head generation.}
As shown in Fig.~\ref{fig:aud_res}, CatPortrait produces the most visually pleasing results among the compared methods, with sharp facial details, stable identity, and natural expressions.
In particular, our method generates more realistic mouth shapes and preserves better overall facial coherence under diverse speech inputs.
By contrast, AniPortrait tends to produce lower-quality facial details, especially around the mouth region, where noticeable artifacts can appear; for example, in the rightmost identity sample, the teeth structure becomes visibly distorted.
Hallo generally achieves reasonable lip motion, but may occasionally produce unnatural mouth shapes.
Hallo2 preserves overall appearance well, yet slight identity drift can still be observed in some cases.
SadTalker often achieves plausible lip synchronization, but its generated expressions are comparatively rigid and less vivid.
These visual comparisons suggest that CatPortrait better balances visual fidelity, speech-driven motion, and identity consistency by leveraging structured 3D facial dynamics together with the pretrained video prior.

\begin{figure}
    \centering
    \includegraphics[width=1\linewidth]{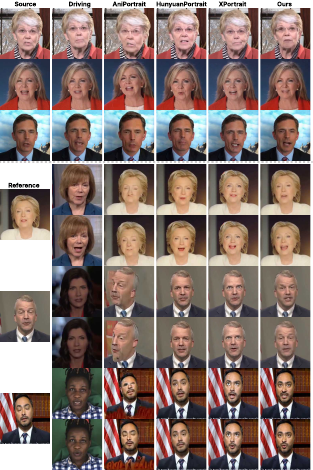}
    \caption{Qualitative comparison on video-driven portrait reenactment, including self-reenactment and cross-reenactment. 
    }
    \label{fig:self_re}
\end{figure}

\noindent\textbf{Video-driven talking head generation.}
As shown in Fig.~\ref{fig:self_re}, CatPortrait delivers strong qualitative results in both self-reenactment and cross-reenactment. In the upper part of the figure, which corresponds to self-reenactment, CatPortrait accurately follows the driving frames in terms of facial expression, mouth motion, and head pose, while maintaining temporally coherent and photorealistic results. HunyuanPortrait and X-Portrait also produce strong reenactment results overall. However, HunyuanPortrait may exhibit mismatched lip motion in certain frames, where the generated mouth shape does not fully align with the driving expression, while AniPortrait shows more noticeable identity drift.
In the lower part of the figure, which corresponds to cross-reenactment, CatPortrait remains robust even when motion and identity come from different subjects. Our method effectively transfers the facial motion and expression style of the driving video while preserving the target identity and maintaining natural facial appearance. X-Portrait sometimes produces less natural eye regions, leading to visible artifacts. HunyuanPortrait performs strongly overall, but CatPortrait achieves comparably natural or better-looking results while preserving more faithful motion cues in challenging cross-identity cases. In contrast, AniPortrait is more prone to visual collapse and unstable facial structures. These observations are consistent with the quantitative results and further demonstrate the advantage of using disentangled 3D face guidance as an explicit intermediate control signal for robust video-driven portrait reenactment.

\subsection{Ablation Study}
We conduct ablation studies only on the video-driven portrait reenactment task, since this setting directly reflects the effectiveness of the proposed in-context 3D guidance for facial motion transfer and identity preservation. 

Specifically, we study four aspects of the proposed framework. First, we investigate different spatial concatenation layouts, including left-right and top-bottom concatenation with different ordering of the RGB portrait input and the rendered structural guidance. Second, we replace the 3DMM-based guidance with 2D landmarks to examine whether the disentangled 3D facial representation is necessary. Third, we remove the identity replacement step at inference time, i.e., the rendered guidance is driven by the source identity instead of replacing the identity coefficients with those extracted from the reference portrait, in order to verify the role of identity-motion disentanglement in cross-identity reenactment. Fourth, we compare full-parameter fine-tuning and LoRA-based fine-tuning to analyze whether the proposed framework can be adapted in a parameter-efficient manner.

As shown in Table~\ref{tab:ablation_video_driven}, the proposed default setting, which places the RGB portrait on the left and the rendered 3DMM guidance on the right, achieves the best overall trade-off between motion fidelity, temporal coherence, and identity preservation. Changing the concatenation order or replacing left-right concatenation with top-bottom concatenation leads to only limited performance variation, suggesting that the proposed framework is not overly sensitive to the specific spatial arrangement under the current backbone. Nevertheless, the default left-right layout remains the most stable setting overall and is therefore adopted in all other experiments.

\begin{table*}[t]
\centering
\caption{Ablation study on video-driven portrait reenactment. All variants are evaluated under the same protocol. Lower is better for FID, FVD, LMD, LPIPS, AED, and AKD. Higher is better for ID Sim. and Image Quality.}
\label{tab:ablation_video_driven}
\resizebox{\textwidth}{!}{
\begin{tabular}{lcccccccc}
\toprule
\multirow{2}{*}{Method Variant} 
& \multicolumn{4}{c}{Self-Reenactment} 
& \multicolumn{4}{c}{Cross-Reenactment} \\
\cmidrule(lr){2-5}\cmidrule(lr){6-9}
& FID$\downarrow$ & FVD$\downarrow$ & LMD$\downarrow$ & LPIPS$\downarrow$
& AED$\downarrow$ & AKD$\downarrow$ & ID Sim.$\uparrow$ & IQ$\uparrow$ \\
\midrule
Left RGB + Right 3DMM (Full FT)
& 88.79 & 104.99 & 1.503 & 0.0757
& 0.9563 & 6.9923 & 0.8025 & 37.5668 \\

Left 3DMM + Right RGB
& 88.84 & 105.35 & 1.497 & 0.0769
& 0.9688 & 7.0831 & 0.7989 & 37.4215 \\

Top RGB + Bottom 3DMM
& 90.21 & 107.43 & 1.524 & 0.0774
& 0.9736 & 7.1205 & 0.7967 & 37.3084 \\

Top 3DMM + Bottom RGB
& 90.87 & 107.96 & 1.533 & 0.0782
& 0.9798 & 7.1886 & 0.7941 & 37.1862 \\

\midrule
2D Landmark guidance
& 95.84 & 114.72 & 1.621 & 0.0829
& 1.0476 & 7.5429 & 0.7794 & 36.4812 \\

w/o ID replacement
& 88.81 & 104.84 & 1.505 & 0.0749
& 1.1268 & 7.9342 & 0.7426 & 37.1024 \\

\midrule
Left RGB + Right 3DMM (LoRA)
& 91.46 & 108.57 & 1.547 & 0.0783
& 0.9914 & 7.2148 & 0.7913 & 37.0417 \\
\bottomrule
\end{tabular}}
\end{table*}

Replacing 3DMM guidance with 2D landmarks leads to consistent degradation on both self-reenactment and cross-reenactment. This result indicates that sparse 2D structural cues are insufficient to provide the same level of controllable motion guidance as the disentangled 3D facial representation. In particular, the performance drop on AED, AKD, and ID Similarity suggests that 3DMM is more effective in decoupling facial motion from identity and recomposing motion onto the target portrait.

Removing the identity replacement step mainly harms cross-identity reenactment. Although the model can still follow the driving motion to some extent, identity preservation deteriorates noticeably when the rendered guidance keeps the source identity instead of the reference identity. This observation verifies that replacing the identity coefficients in the rendered white-mesh guidance is important for preserving target identity under cross-identity motion transfer.

Compared with full-parameter fine-tuning, LoRA-based fine-tuning retains competitive performance overall, indicating that the proposed CatPortrait framework can also be adapted in a parameter-efficient manner. However, full fine-tuning still tends to provide slightly stronger motion fidelity and visual quality, especially in challenging cross-identity cases.

\section{Conclusion}
\label{conclusion}
We present CatPortrait, a simple framework for controllable portrait animation that adapts a pretrained video foundation model without modifying its architecture. By representing facial dynamics with a disentangled 3D face model and injecting texture-free 3D guidance through spatial concatenation, CatPortrait enables controllable facial animation while preserving the pretrained generative prior. Experiments show that it achieves competitive performance in both audio-driven and video-driven settings. We hope this work highlights the potential of 3D guidance as an in-context control signal for pretrained video generators.

\bibliographystyle{splncs04}
\bibliography{main}
\end{document}